\documentclass[runningheads,a4paper]{llncs}

\usepackage[T1]{fontenc}
\usepackage{amssymb}
\usepackage{dsfont}
\usepackage{algorithm2e}
\usepackage{amsmath}
\usepackage{graphicx}

\allowdisplaybreaks

\newtheorem{thm}{Theorem}

\newtheorem{rmk}{Remark}

\allowdisplaybreaks

\title{Bayesian methods for low-rank matrix estimation: short survey and theoretical study}

\titlerunning{Bayesian low-rank matrix estimation}

\author{Pierre Alquier}
\institute{CREST, ENSAE, Universit\'e Paris Saclay
\\
\email{pierre.alquier@ensae.fr}
}
\date{\today}

\authorrunning{Pierre Alquier}

\toctitle{Bayesian methods for low-rank matrix estimation: short survey and theoretical study}
\tocauthor{Pierre Alquier}

\begin{document}

\maketitle

\begin{abstract}

{\bf This is the corrected version of a paper that was published as}

\noindent {\it P. Alquier, Bayesian methods for low-rank matrix estimation: short survey and theoretical study, Algorithmic Learning Theory, 2013, S. Jain, R. Munos, F. Stephan and T. Zeugmann Eds., Springer - Lecture Notes in Artificial Intelligence n. 8139, pp. 309-323.}

\noindent {\bf Since then, a mistake was found in the proofs. We fixed the mistake at the price of a minor change in the final result.~\footnote{We thank Arnak Dalalyan (ENSAE) who found the mistake.}~\footnote{At the time of the original publication, the author was a lecturer at UCD Dublin (School of Mathematical Sciences).}}
\\
\\
The problem of low-rank matrix estimation recently received a lot of attention due to
challenging applications. A lot of work has been done on rank-penalized methods~\cite{Bunea}
and convex relaxation~\cite{Candes2}, both on the theoretical and
applied sides. However, only a few papers considered Bayesian estimation.
In this paper, we review the different type of priors considered on matrices to favour
low-rank. We also prove that the obtained Bayesian estimators, under suitable
assumptions, enjoys the same optimality properties as the ones based on penalization.
\\
\\
{\bf Keywords:} Bayesian inference, collaborative filtering, reduced-rank regression,
matrix completion, PAC-Bayesian bounds, oracle inequalities.
\end{abstract}

\section{Introduction}

The problem of low-rank matrix estimation recently received a lot of attention, 
due to challenging high-dimensional applications provided by recommender systems,
see e.g. the NetFlix challenge~\cite{Bennett}.
Depending on the application, several different models are studied: matrix
completion~\cite{Candes2}, reduced-rank
regression~\cite{Reinsel}, trace regression, e.g.~\cite{KLT}, quantum tomogaphy,
e.g.~\cite{AlquierQuantum}, etc.

In all the above mentionned papers, the
authors considered estimators obtained by minimizing a criterion that is the sum of two
terms: a measure of the quality of data fitting, and a penalization term that is added to avoid
overfitting. This term is usually the rank of the matrix, as in~\cite{Bunea}, or, for
computational reasons, the nuclear norm of the matrix, as in~\cite{Candes2}
(the nuclear norm is the sum of the absolute values of the
singular values, it can be seen as a matrix equivalent of the vectors $\ell_1$ norm). However,
it is to be noted that only a few papers considered Bayesian methods: we mention~\cite{Geweke}
for a first study of reduced-rank regression in Bayesian econometrics,
and more recently~\cite{Lim,Lawrence,Salakhutdinov1} for matrix completion and reduced-rank
regression (a more exhaustive bibliography is given below).

The objective of this paper is twofold: first, in Section~\ref{section_review} we provide a
short survey of the priors
that have been effectively used in various problems of low-rank estimation. We focus
on two models, matrix completion and reduced rank-regression, but all the priors can
be used in any model involving low-rank matrix estimation.

Then, in Section~\ref{section_theorem} we prove a theoretical result on the
Bayesian estimator in the context of reduced rank regression. It should be noted that
for some appropriate choice of the hyperparameters, the rate of convergence is the same as
for penalized methods, up to $\log$ terms. The theoretical study in the context of
matrix completion will be the object of a future work.

\section{Model and priors}
\label{section_review}

In this section, we briefly introduce two models: reduced rank
regression,~\ref{subsection_regression}, and matrix completion,~\ref{subsection_completion}.
We then review the priors used in these models,~\ref{subsection_survey}.

\subsection{Reduced rank regression}
\label{subsection_regression}

In the matrix regression model, we observe two matrices $X$ and $Y$ with
$$ Y =  X B + \mathcal{E} $$
where $X$ is an $\ell\times p$ deterministic matrix, $B$ is a $p\times m$ deterministic
matrix and $\mathcal{E}$ is an $\ell\times m$ random matrix with $\mathbb{E}(\mathcal{E})=0$.
The objective is to estimate the parameter matrix $B$.
This model is sometimes refered as multivariate linear regression, matrix regression
or multitask learning. In many applications, it makes sense to assume that the
matrix $B$ has low rank, i.e. ${\rm rank}(B)\ll \min(p,m)$. In this case,
the model is known as {\it reduced rank regression}, and was studied as early as
\cite{Anderson,Izenman}. We refer the reader to the monograph~\cite{Reinsel} for a
complete introduction.

Depending on the application the authors have in mind, additional assumptions on
the distribution of the noise matrix $\mathcal{E}$ are used:
\begin{itemize}
 \item the entries $\mathcal{E}_{i,j}$ of $\mathcal{E}$ are i.i.d., and the probability
 distribution of $\mathcal{E}_{1,1}$ is bounded, sub-Gaussian or Gaussian
 $\mathcal{N}(0,\sigma^2)$. In this case, note that the likelihood of any matrix $\beta$
 is given by
 $$ \mathcal{L}(\beta|Y,\sigma)
    \propto \exp\left\{ - \frac{1}{2\sigma^2}\|Y-X\beta\|^2_F\right\} $$
 where we let $\|M\|_F$ denote the Frobenius norm, $\|M\|_F^2
= {\rm Tr}(M^T M)$.
 \item as a generalization of the latter case, it is often assumed in econometrics
 papers that the rows $\mathcal{E}_i$ of $\mathcal{E}$ are i.i.d. $\mathcal{N}_m(0,\Sigma)$
 for some $m\times m$ variance-covariance matrix $\Sigma$.
\end{itemize}

In order to estimate $B$, we have to specify a prior on $B$ and, depending on the
assumptions on $\mathcal{E}$, a prior on $\sigma$ or on $\Sigma$. Note however that in most
theoretical papers, it is assumed that $\sigma$ is known, or can be upper bounded,
as in~\cite{Bunea}. This assumption is clearly a limitation but it makes sense in some
applications: see e.g.~\cite{AlquierQuantum} for quantum tomography (that can bee seen as a
special case of reduced rank regression).

In non-Bayesian studies, the estimator considered is usually obtained by minimizing
the least-square criterion $\|Y-XB\|^2_F$ penalized by the rank of the matrix~\cite{Bunea}
or the nuclear norm~\cite{Yuan}.
In~\cite{Bunea}, the estimator $\hat{B}$ obtained by
this method is shown to satisfy, for some constant $\mathcal{C}>0$,
$$ \mathbb{E}(\|X\hat{B}-XB\|_F^2) \leq \mathcal{C}
  \sigma^2 {\rm rank}(B) ({\rm rank}(X)+m) $$
(Corollary 6 p. 1290).

\subsection{Matrix completion}
\label{subsection_completion}

In the problem of matrix completion, one observes
entries $Y_{i,j}$ of an $\ell\times m$ matrix $Y=B+\mathcal{E}$
for $(i,j)$ in a given set of indices $I$. Here again, the noise matrix
satisfies $\mathbb{E}(\mathcal{E})=0$ and the objective is to recover $B$ under the
assumption that ${\rm rank}(B)\ll \min(\ell,m)$. Note that under the assumption that
the $\mathcal{E}_{i,j}$ are i.i.d. $\mathcal{N}(0,\sigma^2)$, the likelihood is given
by
$$ \mathcal{L}(\beta|Y,\sigma) \propto \exp\left\{ - \frac{1}{2\sigma^2}
\sum_{(i,j)\in I}(Y_{i,j}-\beta_{i,j})^2 \right\}. $$
In~\cite{Candes2}, this problem is studied without noise (i.e. $\mathcal{E}=0$), the
general case is studied among others in~\cite{Candes1,Candes3,Gross}.

Note that recently, some authors studied the {\it trace regression model},
that includes linear regression, reduced-rank regression and matrix completion as special
cases: see~\cite{Rohde,Klopp,KLT,KoltStF}. Up to our knowledge, this model has not been
considered from a Bayesian perspective until now, so we will mainly focus on reduced
regression and matrix completion in this paper. However, all the
priors defined for reduced-rank regression can also be used for the more general trace
regression setting.

\subsection{Priors on (approximately) low-rank matrices}
\label{subsection_survey}

It appears that some econometrics models can actually be seen as special cases of
the reduced rank regression. Some of them were studied from a Bayesian perspetive
from the seventies, to our knowledge, it was the first Bayesian study of a reduced
rank regression:
\begin{itemize}
 \item incomplete simultaneous equation model: \cite{Dreze1,Dreze2,Zellner,Kleibergen2},
 \item cointegration: \cite{Bauwens,Kleibergen1,Kleibergen3}.
\end{itemize}
The first systematic treatment of the reduced rank model from a Bayesian
perspective was carried out in \cite{Geweke}. The idea of this paper is to write
the matrix parameter $B$ as $B=MN^T$ for two matrices $M$ and $N$ respectively $p\times k$ and
$m \times k$, and to give a prior on $M$ and $N$ rather than on $B$. Note that the
rank of $B$ is in any case smaller than $k$. So, to choose $k \ll \min(m,p)$ imposes
a low rank structure to the matrix $B$.

The prior in \cite{Geweke} is given by
$$ \pi(M,N,\Sigma) = \pi(M,N)\pi(\Sigma) $$
where $\pi(M,N)$ is a Gaussian shrinkage on all the entries of the matrices:
$$
\pi(M,N)  \propto \exp\left\{-\frac{\tau^2}{2}\left(\|M\|_F^2
           + \|N\|_F^2\right) \right\}
           $$
           for some parameter $\tau>0$.
Then, $\pi(\Sigma)$ is an $\ell$-dimensional inverse-Wishart distribution
with $d$ degrees of freedom and matrix parameter $S$, $\Sigma^{-1}\sim
\mathcal{W}_{\ell}(d,S)$:
$$ \pi(\Sigma) \propto
|\Sigma|^{-\frac{m+d+1}{2}} \exp\left(-\frac{1}{2}{\rm Tr}(S\Sigma^{-1})
\right) .$$
Remark that this prior is particularly convenient as it is then possible
to give explicit forms for the marginal posteriors. This allows an implementation of
the Gibbs algorithm to sample from the posterior. As the formulas are a bit cumbersome,
we do not provide them here, however, the interested reader can find them in \cite{Geweke}.

The weak point in this approach is that the question of the choice of the
reduced rank $k$ is not addressed. It is possible to estimate $M$ and $N$ for any
possible $k$ and to use Bayes factors for model selection,
as in~\cite{Kleibergen3}. Numerical approximation and assessment of convergence
for this method are provided by~\cite{Corander}.

A more recent approach consists in fixing a large $k$, as $k=\min(p,m)$, and then
in calibrating the prior so that it would naturally favour matrices with rank smaller
than $k$ (or, really close to such matrices). To our knowledge, the first attempt
in this direction is~\cite{Lim}. Note that this paper was actually about matrix
completion rather than reduced rank regression, but once again, all the priors
in this subsection can be used in both settings. Here again, we write $B=MN^T$, and
$$ \pi(M,N) \propto  \exp\left\{-\frac{1}{2}\left(
         \sum_{i=1}^{p}\sum_{j=1}^{k} \frac{M_{i,j}^2}{\sigma^2_j} +
         \sum_{i=1}^{m}\sum_{j=1}^{k} \frac{N_{i,j}^2}{\rho^2_{j}}\right) \right\}.$$
In other words, if we write $M=(M_1|\dots|M_k)$ and $N=(N_1|\dots|N_k)$, then
the $M_j$ and $N_j$ are independent and respectively $\mathcal{N}_{p}(0,\sigma^2_j I_p)$
and $\mathcal{N}_{m}(0,\rho^2_j I_m)$ where $I_d$ is the indentity matrix of size $d$.
In order to understand the idea behind this prior, assume for one moment that
$\sigma^2_j$ and $\rho^2_j$ are large for $1\leq j \leq k_0$ and very small for $j>k_0$.
Then, for $j>k_0$, $M_j$ and $N_j$ have entries close to $0$, and so $M_j N_j^T \simeq 0$.
So, the matrix
$$ B = MN^T = \sum_{j=1}^{k} M_j N_j^T \simeq \sum_{j=1}^{k_0} M_j N_j^T ,$$
a matrix that has a rank at most $k_0$.
In practice, the choice of the $\sigma^2_j$'s and $\rho_j^2$'s
is  the main difficulty of this approach. Based on a heuristic, the authors proposed an
estimation of these quantities
that seems to perform well in practice. Remark that the authors assume that
the $\mathcal{E}_{i,j}$ are independent $\mathcal{N}(0,\sigma^2)$ and the parameter $\sigma^2$
is not modelled in the prior (but is still estimated on the data). They finally
propose a variational Bayes approach to approximate the posterior.

Very similar priors were used by~\cite{Lawrence} and in the PMF method
(Probabilistic Matrix Factorisation)
of~\cite{Salakhutdinov1}. However, improved versions were proposed
in~\cite{Salakhutdinov2,Zhou,Babacan,Paisley}:
the authors proposed a full Bayesian treatment of the problem by putting priors
on the hyperparameters. We describe more precisely the prior in~\cite{Salakhutdinov2}:
the $M_j$ and $N_j$ are independent and respectively $\mathcal{N}_{p}(\mu_M,\Sigma_M)$
and $\mathcal{N}_{m}(\mu_N,\Sigma_N)$, and then: $\mu_M\sim\mathcal{N}_p(\mu_0,\beta_0^{-1}
\Sigma_M)$, $\mu_M\sim\mathcal{N}_p(\mu_0,\beta_0^{-1}
\Sigma_N)$,
and finally $\Sigma_M^{-1},\Sigma_N^{-1}\sim\mathcal{W}_{p}(d,S)$. Here again, the
hyperparameters
$\beta_0$, $d$ and $S$ are to be specified. The priors in~\cite{Zhou,Babacan} are quite
similar, and we give more details about the one in~\cite{Babacan} in
Section~\ref{section_theorem}.
In~\cite{Salakhutdinov1,Salakhutdinov2,Zhou,Babacan}, the authors simulate from the
posterior thanks to the Gibbs
sampler (the posterior conditional distribution are explicitely provided e.g.
in~\cite{Salakhutdinov2}). Alternatively,~\cite{Lawrence} uses a stochastic gradient descent
to approximate the MAP (maximum a posteriori).

Some papers proposed a kernelized version of the reduced rank regression and matrix
completion models. Let
$M^{i}$ denote the $i$-th row of $M$ and $N^{h}$ the $h$-th row of $N$. Then, $B=MN^T$ leads
to $B_{i,h} = M^{i} (N^{h})^T $. We can replace this relation by
$$ B_{i,h} = K(M^i,N^h) $$
for some RKHS Kernel $K$. In~\cite{Yu1}, the authors propose a Bayesian formulation
of this model: $B$ is seen as a Gaussian
process on $\{1,\dots,p\}\times\{1,\dots,m\}$ with expectation zero and covariance
function related to the kernel $K$. The same idea is refined in~\cite{Yu2} and applied
successfully to very large datasets, including the NetFlix challenge dataset, thanks to
two algorithms: the Gibbs sampler, and the EM algorithm to approximate the MAP.

Finally, we want to mention the nice theoretical work \cite{Aoyagi1,Aoyagi2}: in these papers,
the authors study the asymptotic performance of Bayesian estimators in the reduced rank
regression model under a general prior $\pi(M,N)$ that has a compactly supported and infinitely
differentiable density. Clearly, the priors aforementioned do not fit the compact
support assumption. The question wether algorithmically tractable priors fit this assumption
is, to our knowledge, still open. In Section~\ref{section_theorem}, we propose a
non-asymptotic analysis of the prior of \cite{Babacan}.

\section{Theoretical analysis}
\label{section_theorem}

In this section, we provide a theoretical analysis of the Bayesian estimators obtained
by using the idea of hierarchical priors of ~\cite{Salakhutdinov2,Zhou,Babacan,Paisley}.
More precisely, we use exactly the prior of~\cite{Babacan} and provide a theoretical
result on the performance of the estimator in the reduced-rank regression model.

Several approaches are available to study the performance of Bayesian estimators:
the asymptotic approach based on Bernstein-von-Mises type theorems, see Chapter 10
in~\cite{vdv}, and a non-asmptotic approach based on PAC-Bayesian inequalities.
PAC-Bayesian inequalities were introduced for classification by~\cite{STW,McAllester}
but tighter bounds and extentions to regression estimation can
be found in~\cite{Catoni2004,Catoni2007,DT1,DT12a,DalSal}. In all approaches, the
variance of the noise is assumed to be known or at least upper-bounded by a given
constant, so we use this framework here. To our knowledge, this is the first application
of PAC-Bayesian bounds to a matrix estimation problem.

\subsection{Theorem}

Following~\cite{Babacan} we write $B=MN^T$ where $M$ is $p\times k$, $N$ is $m \times k$,
$k\leq\min(p,m)$ and then
$$ \pi(M,N|\Gamma) \propto \exp\left[-\frac{1}{2}\left({\rm Tr}(M^T \Gamma^{-1} M)
+{\rm Tr}(N^T \Gamma^{-1} N)\right)
\right] $$
for some diagonal matrix
$$\Gamma = \left(
\begin{array}{c c c}
 \gamma_1 &  \dots & 0 \\ \vdots & \ddots & \vdots \\ 0 & \dots & \gamma_k
\end{array}
\right),$$
the $\gamma_j$ are i.i.d. and $1/\gamma_j \sim {\rm Gamma}(a,b)$:
$$ \pi(M,N) = \int \pi(M,N|\Gamma)\pi(\Gamma) {\rm d}\Gamma$$
where
$$
       \pi(\Gamma) = \frac{b^{ka}}{\Gamma(a)^k} \prod_{j=1}^{k} \left\{ \gamma_j^{-a-1}
              \exp\left(-\frac{b}{\gamma_j}\right)\right\}.$$
We will make one of the following assumptions on the noise:
\begin{itemize}
\item {\bf Assumption (A1):} the entries $\mathcal{E}_{i,j}$ of $\mathcal{E}$ are i.i.d.
           $\mathcal{N}(0,\sigma^2)$, and we know an upper bound $s^2$ for $\sigma^2$.
\item {\bf Assumption (A2):} the entries of $\mathcal{E}$ are iid according to any distribution
supported by
the compact interval $[-\zeta,\zeta]$ with a density $f$ w.r.t. the Lebesgue measure
and $f(x)\geq f_{\min} >0$, and we know an upper bound
$s^2 \geq\mathbb{E}(|\mathcal{E}_{1,1}|)/(2 f_{\min})$.
\end{itemize}
Note that {\bf (A1)} and {\bf (A2)} are special case of the one in \cite{DT1}, the interested
reader can replace these assumptions by the more technical condition given in \cite{DT1}.
We define
$$ \hat{B}_{\lambda} = \int M N^T \hat{\rho}_{\lambda}({\rm d}(M,N)) $$
where $\hat{\rho}_{\lambda}$ is the probability distribution given by
$$ \rho({\rm d} (M,N)) \propto \exp\left( - \lambda \|Y-XMN^T\|_F^2 \right)
\pi({\rm d} (M,N)) .$$
Note that in the case where the entries of $\mathcal{E}$ are i.i.d. $\mathcal{N}(0,\sigma^2)$
then this is the Bayesian posterior, $\rho({\rm d}(M,N))=\pi({\rm d}(M,N)|Y)$, when
$\lambda = 1/(2\sigma^2)$, and so $\hat{B}_{\lambda}$ is the expectation under
the posterior. However, for theoretical reasons, we have to consider
slightly smaller $\lambda$ to prove theoretical results.

\begin{thm}
\label{thm}
Assume that either {\bf (A1)} or {\bf (A2)} is satisfied.
Let us put $a=1$ and $b=\frac{s^2}{2 \ell p k^2 (m^2+p^2)}$ in the prior $\pi(\Gamma)$.
For $\lambda = \frac{1}{4s^2}$ we have, for any $\epsilon>0$,
\begin{multline*}
\mathbb{E}\left(\|X\hat{B}_{\lambda}-XB\|_{F}^2\right)
\leq \inf_{\begin{array}{c}J,M,N \\
M_j,N_j = 0 \text{ when } j\notin J\end{array}} \Biggl\{
   (1+\epsilon) \|X(MN^T-B)\|_{F}^2
    \\ 
    + 6s^2(m+p)|J|\log \left(\frac{1.34\ell p}{s^2}\right)
     + 8s^2 k\log\left(\frac{22.17 \ell p k^2 (m^2+p^2)}{s^2}\right)
   \\
   + \frac{\left(4+\frac{4}{\epsilon}\right) s^2}{\ell p}\left\{ 4s^2 + \|X\|_F^2  \left[ \|N\|_F^2 +
       \|M\|_{F}^2+\frac{s^2}{\ell p}  \right]\right\}
       \\
       + 8s^2 \left(\|N\|_F^2 + \|M\|_{F}^2 + \log(2)\right)
\Biggr\}.
\end{multline*}
\end{thm}

\begin{rmk}
 Note that when all the entries of $X$ satisfy $|X_{i,j}| \leq C$ for some $C>0$,
$\|X\|_F^2 / (\ell p) \leq C^2$. Moreover, let us assume that
 ${\rm rank}(B)=k_0$ and that we can write $B=M N^T$
 with $M_{k_0+1}=\dots = M_k =0$ and $N_{k_0+1}=\dots = N_k = 0$ and $|N_{i,j}|,|M_{i,j}|\leq c$.
 Assume that the noise is Gaussian. Taking for example $\epsilon=4$ we get
\begin{multline*}
\mathbb{E}\left(\|X\hat{B}_{\lambda}-XB\|_{F}^2\right)
\leq 
     50 s^2(m+p)k_0 \Biggl\{ \log(\ell(p\vee m))
     \\ + \log\left(\frac{1}{s^2}\vee 1\right)
                         + 1 + c^2 C^2 + s^2(1+c^2+C^2) \Biggr\}
\end{multline*}
where we remind that $p\vee m = \max(p,m)$.
When ${\rm rank}(X)=p$, we can see that we recover the same upper bound as
in~\cite{Bunea}, up to a $\log(\ell (p\vee m))$ term. This rate (without the $\log$) is known
to be optimal, see~\cite{Bunea} remark (ii) p. 1293 and~\cite{Rohde}.
However, the presence of the terms $\|M\|_F^2$ and $\|N\|_F^2$ can lead to
suboptimal rates in less classical asymptotics where $\|B\|_F$ would grow with the sample size
$\ell$. In the case of linear regression, a way to avoid these terms is to use
heavy-tailed priors as in~\cite{DT1,DT12a}, or compactly supported priors as in~\cite{AL}.
However, it is not clear whether this approach would lead to feasible algorithms
in matrix estimation problems. This question will be the object of a future work.
\end{rmk}

\begin{rmk}
 We do not claim that the choice $b=\frac{s^2}{2 \ell p k^2 (m^2+p^2)}$ is optimal in
 practice. However, from the proof it is clear that our technique requires that $b$
 decreases with the dimension of $B$ as well as with the sample size to produce
 a meaningfull bound. Note that in~\cite{Babacan}, there is no theoretical approach
 for the choice of $b$, but their simulation study tends to show that $b$ must be very
 small for $MN^T$ to be approximately low-rank.
\end{rmk}

\begin{rmk}
In all the above mentionned papers on PAC-Bayesian bounds, it is assumed that the
variance of the noise is known, or upper-bounded by a known constant.
More recently,~\cite{AudibertCatoni} managed to prove PAC-Bayesian
inequalities for regression with unknown variance. However, the approach is rather
involved and it is not clear whether it can be used in our context. This question
will also be addressed in a future work.
\end{rmk}

\subsection{Proof}

First, we state the following result:

\begin{thm}
\label{thmDT1}
Under {\bf (A1)} or {\bf (A2)}, for any $\lambda\leq 1/(4s^2)$, we have
\begin{equation*}
\mathbb{E}\left(\|X\hat{B}_{\lambda}-XB\|_{F}^2\right)
\leq \inf_{\rho} \left\{
\int \|X\mu\nu^T-XB\|_{F}^2 \rho({\rm d}(\mu,\nu)) + \frac{\mathcal{K}(\rho,\pi)}{\lambda}
\right\}
\end{equation*}
where $\mathcal{K}(\rho,\pi)$ stands for the Kullback divergence between $\rho$ and
$\pi$, $\mathcal{K}(\rho,\pi)=\int \log(\frac{{\rm d}\rho}{{\rm d}\pi}){\rm d}\rho$ if
$\rho$ is absolutely continuous with respect to $\pi$ and $\mathcal{K}(\rho,\pi)=\infty$
otherwise.
\end{thm}

\noindent \textit{Proof of Theorem~\ref{thmDT1}.}
 Follow the proof of Theorem 1 in~\cite{DT1} and check that every step is valid when
 $B$ is a matrix instead of a vector.
$\square$

We are now ready to prove our main result.

\noindent \textit{Proof of Theorem~\ref{thm}.}
Let us introduce, for any $c>0$,
the probability distribution $\rho_{M,N,c}({\rm d}\mu,{\rm d}\nu)
\propto \mathbf{1}(\|\mu-M\|_{F}\leq c,\|\nu-N\|_{F} \leq c) \pi({\rm d}\mu,{\rm d}\nu)$.
According to Theorem~\ref{thmDT1} we have
\begin{multline}
\label{stepDT}
\mathbb{E}\left(\|X\hat{B}_{\lambda}-XB\|_{F}^2\right)
\\
\leq \inf_{M,N,c} \left\{
\int \|X\mu\nu^T-XB\|_{F}^2 \rho_{M,N,c}({\rm d}\mu,{\rm d}\nu)
+ \frac{\mathcal{K}(\rho_{M,N,c},\pi)}{\lambda}
\right\}.
\end{multline}
Let us fix $c$, $M$ and $N$.
The remaining steps of the proof are to upper-bound the two terms in the r.h.s.
Both upper bounds will depend on $c$, we will optimize on $c$ after these steps to
end the proof. We have, for any $\epsilon>0$,
\begin{align}
\nonumber
 & \int \|X\mu\nu^T-XB\|_{F}^2 \rho_{M,N,c}({\rm d}\mu,{\rm d}\nu) \\
 \nonumber
 & \hspace{0.5cm} =
 \int \|X\mu\nu^T-XM\nu^T + XM\nu^T-XMN^T
 \\
  \nonumber
  & \hspace{2cm}+XMN^T-XB\|_{F}^2 \rho_{M,N,c}({\rm d}\mu,{\rm d}\nu)
\\
 \nonumber
 & \hspace{0.5cm} \leq
 \left(1+\frac{1}{\epsilon}\right)  \int \|X\mu\nu^T-XM\nu^T + XM\nu^T-XMN^T\|_F^2 \rho_{M,N,c}({\rm d}\mu,{\rm d}\nu)
 \\
  \nonumber
  & \hspace{2cm}+ \left(1+\epsilon\right) \int \|XMN^T-XB\|_{F}^2 \rho_{M,N,c}({\rm d}\mu,{\rm d}\nu)
\\
  \nonumber
 & \hspace{0.5cm} \leq  \left(2+\frac{2}{\epsilon}\right)  \int \Bigl\{ \|X\mu\nu^T-XM\nu^T\|_{F}^2
     \\
      \nonumber
     & \hspace{2cm}
    + \|XM\nu^T-XMN^T\|_{F}^2
  \Bigr\} \rho_{M,N,c}({\rm d}\mu,{\rm d}\nu)
       \\
      \nonumber
     & \hspace{2cm}
+\left(1+\epsilon\right) \|XMN^T-XB\|_{F}^2 \\
      \nonumber
 & \hspace{0.5cm} \leq   \left(2+\frac{2}{\epsilon}\right) \|X\|_F^2  \int \left\{ \|\mu-M\|_{F}^2\|\nu\|_F^2 +
             \|M\|_{F}^2\|\nu-N\|_{F}^2
  \right\} \rho_{M,N,c}({\rm d}\mu,{\rm d}\nu)
         \\
          \nonumber
     & \hspace{2cm} + \left(1+\epsilon\right) \|XMN^T-XB\|_{F}^2
     \\
 & \hspace{0.5cm} \leq   \left(4+\frac{4}{\epsilon}\right)  c^2\|X\|_F^2  \left\{ (\|N\|_F^2+c^2) + 
             \|M\|_{F}^2\right\}
             \\
             \label{step1stterm}
    & \hspace{2cm} + \left(1+\epsilon\right) \|XMN^T-XB\|_{F}^2.
\end{align}

Now, we deal with the second term:
$$
\mathcal{K}(\rho_{M,N,c},\pi)
 = \log \frac{1}{\pi(\{\mu,\nu:\|\mu-M\|_{F}\leq c,\|\nu-N\|_{F} \leq c\})}.
$$
We remind that
$M=(M_1|\dots|M_k)$ and $N=(N_1|\dots|N_k)$
and let us denote $J$ the subset of $\{1,\dots,k\}$ such that $M_j=N_j=0$ for $j\notin J$.
We let $k_0$ denote the cardinality of $J$, $k_0=|J|$. Note that we have ${\rm rank}(MN^T)
\leq k_0$.
For any $\kappa\in(0,1)$ let $E_{\kappa}$ be the event
$$ \left\{\frac{\kappa}{2}<|\gamma_j| < \kappa \text{ for any } j\notin J \text{ and }
   |\gamma_j-1|< \frac{1}{2} \text{ for any } j\in J \right\} .$$
Then
\begin{align}
\nonumber
\mathcal{K}(\rho_{M,N,c},\pi)
& \leq \log \frac{1}{\int \pi(\{\mu,\nu:\|\mu-M\|_{F}\leq c,\|\nu-N\|_{F} \leq c
 \}|\Gamma)\pi(\Gamma){\rm d}\Gamma}
\\
\nonumber
& =  \log \frac{1}{\int \pi(\{\|\mu-M\|_{F}\leq c\}|\Gamma)\pi(\Gamma){\rm d}\Gamma}
\\
\nonumber
& \hspace{1cm}
+  \log \frac{1}{\int \pi(\{\|\nu-M\|_{F}\leq c\}|\Gamma)\pi(\Gamma){\rm d}\Gamma}
\\
\nonumber
&\leq  \log \frac{1}{\int_{E_\kappa} \pi(\{\|\mu-M\|_{F}\leq c\}|\Gamma)\pi(\Gamma){\rm d}\Gamma}
\\
\label{stepKL}
& \hspace{1cm}
+\log \frac{1}{\int_{E_{\kappa}} \pi(\{\|\nu-M\|_{F}\leq c\}|\Gamma)\pi(\Gamma){\rm d}\Gamma}.
\end{align}
By symmetry, we will only bound the first of these two terms.
We have
\begin{align}
 \nonumber
 & \int_{E_\kappa} \pi(\{\|\mu-M\|_{F}\leq c\}|\Gamma)\pi(\Gamma){\rm d}\Gamma \\
  \nonumber
 & \hspace{0.5cm} =  \int_{E_\kappa} \pi\left( \left.\sum_{i=1}^{p} \sum_{j=1}^{k} (\mu_{i,j}
               -M_{i,j})^2 \leq c^2\right|\Gamma\right)\pi(\Gamma){\rm d}\Gamma \\
                \nonumber
 & \hspace{0.5cm} \geq \int_{E_\kappa} \pi\left(\left.\forall i, \forall j,
   (\mu_{i,j}-M_{i,j})^2
  \leq \frac{c^2}{pk}\right|\Gamma\right)\pi(\Gamma){\rm d}\Gamma \\
   \nonumber
 & \hspace{0.5cm} =\int_{E_\kappa}
 \left\{ 1- \pi\left(\left.\exists i\in\{1,\dots,p\}, \exists j\notin J,
   (\mu_{i,j}-M_{i,j})^2
  \geq \frac{c^2}{pk}\right|\Gamma\right)\right\}
  \\
   \nonumber
  & \hspace{2cm} \prod_{i=1}^{p}\prod_{j\in J}
 \pi\left(\left. (\mu_{i,j}-M_{i,j})^2
  \leq \frac{c^2}{pk}\right|\Gamma\right)
  \pi(\Gamma){\rm d}\Gamma \\
   \nonumber
  & \hspace{0.5cm} \geq \int_{E_\kappa}
 \left\{ 1- \sum_{i=1}^{p} \sum_{j\notin J} \pi\left(\left.
   (\mu_{i,j}-M_{i,j})^2
  \geq \frac{c^2}{pk}\right|\Gamma\right)\right\}
  \\
  & \hspace{2cm} \prod_{i=1}^{p}\prod_{j\in J}
 \pi\left(\left. (\mu_{i,j}-M_{i,j})^2
  \leq \frac{c^2}{pk}\right|\Gamma\right)
  \pi(\Gamma){\rm d}\Gamma. \label{stepPi}
\end{align}
We lower-bound the three factors in the integral in~\eqref{stepPi} separately.
First, note that, on $E_{\kappa}$,
\begin{align}
 \nonumber
\pi(\Gamma) & = \prod_{j=1}^{k} \frac{b^{a}}{\Gamma(a)}
                                     \gamma_j^{-a-1} \exp\left(-\frac{b}{\gamma_j}\right) \\
                                      \nonumber
& =  \frac{b^{ka}}{\Gamma(a)^k} \left\{
 \prod_{j \in J} \gamma_j^{-a-1} \exp\left(-\frac{b}{\gamma_j}\right) \right\}
 \left\{
 \prod_{j \notin J} \gamma_j^{-a-1} \exp\left(-\frac{b}{\gamma_j}\right) \right\} \\
  \nonumber
& \geq  \frac{b^{ka}}{\Gamma(a)^k} \left\{ \kappa^{-a-1}
     \exp\left(-\frac{2b}{\kappa}\right) \right\}^{k-k_0}
 \left\{
\left(\frac{3}{2}\right)^{-a-1}
     \exp\left(-2b\right)\right\}^{k_0} \\
      \nonumber
   & \geq  \frac{b^{ka}}{\Gamma(a)^k} \exp\left\{-2b\left(
   \frac{k-k_0}{\kappa}-k
   \right)\right\} \left(\frac{3}{2}\right)^{(-a-1)k_0} \kappa^{(-a-1)(k-k_0)} \\
   \label{stepPi1}
  & \geq \frac{b^{ka}}{\Gamma(a)^k} \left(\frac{2}{3}\right)^{(a+1)k}
     \exp\left\{\frac{-2bk}{\kappa}\right\} \kappa^{(-a-1)(k-k_0)}.
\end{align}
On $E_\kappa$, and for $j\notin J$:
$$
    \pi\left(\left. |\mu_{i,j}|
   \geq \frac{c}{\sqrt{pk}} \right|\Gamma \right)
    = 2\Phi\left(\frac{c}{\sqrt{pk\gamma_j}}\right)
$$
where $\Phi$ is the c.d.f. of $\mathcal{N}(0,1)$. We use the classical inequality
$$ \Phi(x) \leq \frac{\exp\left(-\frac{x^2}{2}\right)}{2} $$
to get:
$$
    \pi\left(\left. |\mu_{i,j}|
   \geq \frac{c}{\sqrt{pk}} \right|\Gamma \right)
    \leq  \exp\left(-\frac{c^2}{2pk\gamma_j}\right)
    \leq \exp\left(-\frac{c^2}{2pk\kappa}\right)
$$
and finally
\begin{equation}
\sum_{i=1}^{p} \sum_{j\notin J}
\pi\left(\left. (\mu_{i,j}-M_{i,j})^2
   \geq \frac{c^2}{pk}\right|\Gamma \right)
    \leq p k_0  \exp\left(-\frac{c^2}{2pk\kappa}\right).
\label{stepPi2}
\end{equation}
Then, on $E_\kappa$, and for $j\in J$:
\begin{align*}
\pi\left( \left. (\mu_{i,j}-M_{i,j})^2
   \leq \frac{c^2}{pk}\right|\Gamma \right)
   & = \pi\left(\left. (\mu_{i,j}-M_{i,j})^2
   \leq \frac{c^2}{pk}\right|\Gamma \right) \\
   & = \frac{1}{\sqrt{2\pi\gamma_j}}
  \int_{M_{i,j}-\frac{c}{\sqrt{pk}}}^{M_{i,j}+\frac{c}{\sqrt{pk}}}
  \exp\left(-\frac{x^2}{2\gamma_j}\right) {\rm d}x \\
  & \geq c \sqrt{\frac{2}{\pi pk\gamma_j}}
              \exp\left(- \frac{M_{i,j}^2}{\gamma_j} - \frac{c^2}{pk\gamma_j} \right) \\
     & \geq c \sqrt{\frac{4}{3\pi pk}}
         \exp\left(- 2M_{i,j}^2-\frac{2 c^2}{pk}\right) 
\end{align*}
and so
\begin{align}
& \prod_{i=1}^{p} \prod_{j\in J}
 \pi\left(\left. (\mu_{i,j}-M_{i,j})^2
\leq \frac{c^2}{pk}\right|\Gamma\right)
 \nonumber
  \\
  & \hspace{1cm} 
\geq  \left(c \sqrt{\frac{4}{3\pi pk}} \right)^{pk_0}
        \exp\left(- 2\|M\|_{F}^2- 2 c^2\right) .
        \label{stepPi3}
\end{align}
We plug~\eqref{stepPi1},~\eqref{stepPi2} and~\eqref{stepPi3} into~\eqref{stepPi}
and we obtain:
\begin{align*}
 &\int_{E_\kappa}
 \pi(\{\|\mu-M\|_{F}\leq c\}|\Gamma)\pi(\Gamma){\rm d}\Gamma
 \\
 & \hspace{1cm}
 \geq \int_{E_\kappa}  \kappa^{(-a-1)(k-k_0)}
 \frac{b^{ka}}{\Gamma(a)^k} \left(\frac{2}{3}\right)^{(a+1)k}
     \exp\left\{\frac{-2bk}{\kappa}\right\}
     \left(c \sqrt{\frac{4}{3\pi pk}} \right)^{pk_0}
    \\
    & \hspace{2cm} \exp\left(- 2\|M\|_{F}^2- 2 c^2\right)
        \left(1- p k_0  \exp\left(-\frac{c^2}{2pk\kappa}\right)\right)
        {\rm d}\gamma_1 \dots {\rm d}\gamma_k
 \\
 &\hspace{1cm} = \left(\frac{\kappa}{2}\right)^{k-k_0}  \kappa^{(-a-1)(k-k_0)}
     \frac{b^{ka}}{\Gamma(a)^k} \left(\frac{2}{3}\right)^{(a+1)k}
     \exp\left\{\frac{-2bk}{\kappa}\right\}
     \left(c \sqrt{\frac{4}{3\pi pk}} \right)^{pk_0}
     \\
      & \hspace{2cm} \exp\left(- 2\|M\|_{F}^2- 2 c^2\right)
        \left(1- p k_0  \exp\left(-\frac{c^2}{2pk\kappa}\right)\right).
\end{align*}
Now, let us impose the following restrictions:
$b=\kappa\leq \frac{c^2}{2pk \log(2pk)}\leq \frac{c^2}{2pk \log(2pk_0)}$
so the last factor is $\geq 1/2$. So we have:
\begin{align*}
 &\int_{E_\kappa} \pi(\{\|\mu-M\|_{F}\leq c\}|\Gamma)\pi(\Gamma){\rm d}\Gamma
 \\
 & \hspace{1cm} \geq
     \frac{\kappa^{ka}}{\Gamma(a)^k} \frac{2^{ak+1}}{3^{(a+1)^k}}
     \exp\left\{-2k\right\}
     \left(c \sqrt{\frac{4}{3\pi pk}} \right)^{pk_0}
     \\
      & \hspace{2cm} \exp\left(- 2\|M\|_{F}^2- 2 c^2\right).
\end{align*}
So,
\begin{multline}
  \label{stepKL1}
 \log\frac{1}{\int_{E_\kappa} \pi(\{\|\mu-M\|_{F}\leq c\}|\Gamma)\pi(\Gamma){\rm d}\Gamma}
   \leq 2c^2 + 2\|M\|_{F}^2 
 \\
+ \log(2)
       +pk_0 \log\left(\frac{1}{c}\sqrt{\frac{3\pi pk}{4}}\right)
        + k \log\left(\frac{\Gamma(a)3^{a+1}\exp(2)}{\kappa^{a+1}2^a} \right).
\end{multline}
By symmetry,
\begin{multline}
 \log\frac{1}{\int_{E_\kappa} \pi(\{\|\nu-N\|_{F}\leq c\}|\Gamma)\pi(\Gamma){\rm d}\Gamma}
  \leq 2c^2 + 2\|N\|_{F}^2 + \log(2)
  \\
 \label{stepKL2}
       +m k_0 \log\left(\frac{1}{c}\sqrt{\frac{3\pi pk}{4}}\right)
        + k \log\left(\frac{\Gamma(a)3^{a+1}\exp(2)}{\kappa^{a+1}2^a} \right),
\end{multline}
and finally, plugging~\eqref{stepKL1} and~\eqref{stepKL2} into~\eqref{stepKL}
\begin{align}
& \mathcal{K}(\rho_{M,N,c},\pi) \leq 4c^2 + 2\|M\|_{F}^2 + 2\|N\|_{F}^2
   + 2\log(2) 
   \nonumber
   \\
   \label{stepKLend}
 & \hspace{1cm}+(m + p)k_0 \log\left(\frac{1}{c}\sqrt{\frac{3\pi pk}{4}}\right)
        + 2 k \log\left(\frac{\Gamma(a)3^{a+1}\exp(2)}{\kappa^{a+1}2^a} \right).
\end{align}
Finally, we can plug~\eqref{step1stterm} and~\eqref{stepKLend} into~\eqref{stepDT}:
\begin{multline*}
\mathbb{E}\left(\|X\hat{B}_{\lambda}-XB\|_{F}^2\right)
\\
\leq \inf_{\begin{array}{c}J,M,N,c \\
M_j,N_j = 0 \text{ when } j\notin J\end{array}} \Biggl\{
\left(4+\frac{4}{\epsilon}\right) c^2\|X\|_F^2  \left\{ \|N\|_F^2 +
             \|M\|_{F}^2+c^2\right\}
    \\
    + (1+\epsilon) \|X(MN^T-B)\|_{F}^2 + \frac{4c^2 + 2\|M\|_{F}^2 + 2\|N\|_{F}^2
   + 2\log(2) }{\lambda}
   \\
   + \frac{(m + p)|J| \log\left(\frac{1}{c}\sqrt{\frac{3\pi pk}{4}}\right)
        + 2 k \log\left(\frac{\Gamma(a)3^{a+1}\exp(2)}{\kappa^{a+1}2^a} \right)}{\lambda}
\Biggr\}.
\end{multline*}

Let us put $c=\sqrt{s^2/\ell p}$ to get:
\begin{multline*}
\mathbb{E}\left(\|X\hat{B}_{\lambda}-XB\|_{F}^2\right)
\leq \inf_{\begin{array}{c}J,M,N \\
M_j,N_j = 0 \text{ when } j\notin J\end{array}} \Biggl\{
   \left(1+\epsilon \right) \|X(MN^T-B)\|_{F}^2
    \\ 
   + \frac{(m + p)|J| \log\left(p
      \sqrt{\frac{\ell k 3\pi}{4 s^2}}\right)
        + 2 k \log\left(\frac{\Gamma(a)3^{a+1}\exp(2)}{\kappa^{a+1}2^a} \right)}{\lambda}
  \\
      + \frac{2\|M\|_{F}^2 + 2\|N\|_{F}^2
   + 2\log(2) }{\lambda}
   \\
   +\frac{ \left(4+\frac{4}{\epsilon}\right) s^2 \left\{ \frac{1}{\lambda} + \|X\|_F^2  \left[ \|N\|_F^2 +
       \|M\|_{F}^2+\frac{s^2}{\ell p}  \right] \right\} }{\ell p}
\Biggr\}.
\end{multline*}
Finally, remember that the conditions of the theorem impose that $a=1$, and
$b=\frac{s^2}{2 \ell p k^2(m^2+p^2)}$. However, we used until now that $b=\kappa$,
that $\kappa<1/2$, that $\kappa \leq c^2/(2pk \log(2pk))
   = s^2/(2p^2 \ell  k \log(2pk))$, and that $\kappa \leq c^2/(2mk \log(2mk))
   = s^2/(2m p \ell  k \log(2mk))$.
Remember that $k\leq \min(p,m)$ so all these equations are compatible. We obtain:
\begin{multline*}
\mathbb{E}\left(\|X\hat{B}_{\lambda}-XB\|_{F}^2\right)
\leq \inf_{\begin{array}{c}J,M,N \\
M_j,N_j = 0 \text{ when } j\notin J\end{array}} \Biggl\{
   \left(1+\epsilon \right) \|X(MN^T-B)\|_{F}^2
   \\
   + \frac{(m + p)|J| \log\left(p
      \sqrt{\frac{\ell k 3\pi}{4 s^2}}\right)
        + 2 k \log\left(\frac{2 \ell p k^2 (m^2+p^2) 3\exp(2)}{s^2}\right)}{\lambda}
  \\
      + \frac{2\|M\|_{F}^2 + 2\|N\|_{F}^2
   + 2\log(2) }{\lambda}
   \\
   +\frac{ \left(4+\frac{4}{\epsilon}\right) s^2 \left\{ \frac{1}{\lambda} + \|X\|_F^2  \left[ \|N\|_F^2 +
       \|M\|_{F}^2+\frac{s^2}{\ell p}  \right] \right\} }{\ell p}
\Biggr\}.
\end{multline*}
This ends the proof. $\square$

\section{Conclusion}

We proved that the use of Gaussian priors in reduced-rank regression models leads
to nearly optimal rates of convergence. As mentionned in the paper, alternative
priors would possibly lead to better bounds but could also result in less computationaly
efficient methods (computational efficiency is a major issue when dealing with
high-dimensional datasets such as the NetFlix dataset). A complete exploration of
this issue will be addressed in future works.

\bibliographystyle{splncs}

\begin{thebibliography}{10}

\bibitem{Bunea}
Bunea, F., She, Y., Wegkamp, M.H.:
\newblock Optimal selection of reduced rank estimators of high-dimensional
  matrices.
\newblock The Annals of Statistics \textbf{39} (2011)  1282--1309

\bibitem{Candes2}
Cand\`es, E., Tao, T.:
\newblock The power of convex relaxation: Near-optimal matrix completion.
\newblock IEEE Transactions on Information Theory \textbf{56} (2009)
  2053--2080

\bibitem{Bennett}
Bennett, J., Lanning, S.:
\newblock The netflix prize.
\newblock In: Proceedings of KDD Cup and Workshop 07. (2007)

\bibitem{Reinsel}
Reinsel, G.C., Velu, R.P.:
\newblock Multivariate reduced-rank regression: theory and applications.
\newblock Springer Lecture Notes in Statistics 136 (1998)

\bibitem{KLT}
Koltchinskii, V., Lounici, K., Tsybakov, A.B.:
\newblock Nuclear-norm penalization and optimal rates for noisy low-rank matrix
  completion.
\newblock The Annals of Statistics \textbf{39} (2011)  2302--2329

\bibitem{AlquierQuantum}
Alquier, P., Butucea, C., Hebiri, M., Meziani, K., Morimae, T.:
\newblock Rank-penalized estimation of a quantum system.
\newblock Preprint arXiv:1206.1711 (2012)

\bibitem{Geweke}
Geweke, J.:
\newblock Bayesian reduced rank regression in econometrics.
\newblock Journal of Econometrics \textbf{75} (1996)  121--146

\bibitem{Lim}
Lim, Y.J., Teh, Y.W.:
\newblock Variational bayesian approach to movie rating prediction.
\newblock In: Proceedings of KDD Cup and Workshop 07. (2007)

\bibitem{Lawrence}
Lawrence, N.D., Urtasun, R.:
\newblock Non-linear matrix factorization with {G}aussian processes.
\newblock In: Proceedings of the 26th annual International Conference on
  Machine Learning (ICML09), ACM, New York (2009)  601--608

\bibitem{Salakhutdinov1}
Salakhutdinov, R., Mnih, A.:
\newblock Bayesian probabilistic matrix factorization.
\newblock In Platt, J.C., Koller, D., Singer, Y., Roweis, S., eds.: Advances in
  Neural Information Processing Systems 20 (NIPS2007), Cambridge, MIT Press
  (2008)

\bibitem{Anderson}
Anderson, T.:
\newblock Estimating linear restrictions on regression coefficients for
  multivariate normal distributions.
\newblock Annals of Mathematical Statistics \textbf{22} (1951)  327--351

\bibitem{Izenman}
Izenman, A.:
\newblock Reduced rank regression for the multivariate linear model.
\newblock Journal of Multivariate Analysis \textbf{5} (1975)  248--264

\bibitem{Yuan}
Yuan, M., Ekici, A., Lu, Z., Monteiro, R.:
\newblock Dimension reduction and coefficient estimation in multivariate linear
  regression.
\newblock Journal of the Royal Statistical Society - Series B \textbf{69}
  (2007)  329--346

\bibitem{Candes1}
Cand\`es, E., Plan, Y.:
\newblock Matrix completion with noise.
\newblock Proceedings of the IEEE \textbf{98} (2009)  625--636

\bibitem{Candes3}
Cand\`es, E., Recht, B.:
\newblock Exact matrix completion via convex optimization.
\newblock Foundations of Computational Mathematics \textbf{9} (2009)  717--772

\bibitem{Gross}
Gross, D.:
\newblock Recovering low-rank matrices from few coefficients in any basis.
\newblock {IEEE} Transactions on Information Theory \textbf{57} (2011)
  1548--1566

\bibitem{Rohde}
Rohde, A., Tsybakov, A.B.:
\newblock Estimation of high-dimensional low-rank matrices.
\newblock The Annals of Statistics \textbf{39} (2011)  887--930

\bibitem{Klopp}
Klopp, O.:
\newblock Rank-penalized estimators for high-dimensionnal matrices.
\newblock Electronic Journal of Statistics \textbf{5} (2011)  1161--1183

\bibitem{KoltStF}
Koltchinskii, V.:
\newblock Oracle Inequalities in Empirical Risk Minimization and Sparse
  Recovery Problems.
\newblock Springer Lecture Notes in Mathematics (2011)

\bibitem{Dreze1}
Dreze, J.H.:
\newblock Bayesian limited information analysis of the simultaneous equation
  model.
\newblock Econometrica \textbf{44} (1976)  1045--1075

\bibitem{Dreze2}
Dreze, J.H., Richard, J.F.:
\newblock Bayesian analysis of simultaneous equation models.
\newblock In Griliches, Z. ans~Intriligater, J.F., ed.: Handbook of
  econometrics, vol. 1, North-Holland, Amsterdam (1983)

\bibitem{Zellner}
Zellner, A., Min, C., Dallaire, D.:
\newblock Bayesian analysis of simultaenous equation and related models using
  the {G}ibbs sampler and convergence checks.
\newblock H. G. B. Alexander Research Founsation working paper, University of
  Chicago (1993)

\bibitem{Kleibergen2}
Kleibergen, F., van Dijk, H.K.:
\newblock Bayesian simultaneous equation analysis using reduced rank
  structures.
\newblock Econometric theory \textbf{14} (1998)  699--744

\bibitem{Bauwens}
Bauwens, L., Lubrano, M.:
\newblock Identification restriction and posterior densities in cointegrated
  gaussian var systems.
\newblock In Fomby, T.M., Carter~Hill, R., eds.: Advances in econometrics, vol.
  11(B), JAI Press, Greenwich (1993)

\bibitem{Kleibergen1}
Kleibergen, F., van Dijk, H.K.:
\newblock On the shape of the likelihood-posterior in cointegration models.
\newblock Econometric theory \textbf{10} (1994)  514--551

\bibitem{Kleibergen3}
Kleibergen, F., Paap, R.:
\newblock Priors, posteriors and bayes factors for a bayesian analysis of
  cointegration.
\newblock Journal of Econometrics \textbf{111} (2002)  223--249

\bibitem{Corander}
Corander, J., Villani, M.:
\newblock Bayesian assessment of dimensionality in reduced rank regression.
\newblock Statistica Neerlandica \textbf{58} (2004)  255--270

\bibitem{Salakhutdinov2}
Salakhutdinov, R., Mnih, A.:
\newblock Bayesian probabilistic matrix factorization using markov chain monte
  carlo.
\newblock In: Proceedings of the 25th annual International Conference on
  Machine Learning (ICML08), ACM, New York (2008)

\bibitem{Zhou}
Zhou, M., Wang, C., Chen, M., Paisley, J., Dunson, D., Carin, L.:
\newblock Nonparametric bayesian matrix completion.
\newblock In: IEEE Sensor Array and Multichannel Signal Processing Workshop.
  (2010)

\bibitem{Babacan}
Babacan, S.D., Luessi, M., Molina, R., Katsaggelos, A.K.:
\newblock Low-rank matrix completion by variational sparse bayesian learning.
\newblock In: IEEE International Conference on Audio, Speech and Signal
  Processing, Prague (Czech Republic) (2011)  2188--2191

\bibitem{Paisley}
Paisley, J., Carin, L.:
\newblock A nonparametric bayesian model for kernel matrix completion.
\newblock In: Proceedings of ICASSP 2010, Dallas, USA. (2010)

\bibitem{Yu1}
Yu, K., Tresp, V., Schwaighofer, A.:
\newblock Learning {G}aussian processes for multiple tasks.
\newblock In: Proceedings of the 22th annual International Conference on
  Machine Learning (ICML05). (2005)

\bibitem{Yu2}
Yu, K., Lafferty, J., Zhu, S., Gong, Y.:
\newblock Large-scale collaborative prediction using a non-parametric random
  effects model.
\newblock In: Proceedings of the 26th annual International Conference on
  Machine Learning (ICML09), ACM, New York (2009)

\bibitem{Aoyagi1}
Aoyagi, M., Watanabe, S.:
\newblock The generalization error of reduced rank regression in bayesian
  estimation.
\newblock In: International Symposium on Information Theory and its
  Applications (ISITA2004), Parma (Italy) (2004)

\bibitem{Aoyagi2}
Aoyagi, M., Watanabe, S.:
\newblock Stochastic complexities of reduced rank regression in bayesian
  estimation.
\newblock Neural networks \textbf{18} (2005)  924--933

\bibitem{vdv}
van~der Vaart, A.W.:
\newblock Asymptotic Statistics.
\newblock Cambridge University Press (1998)

\bibitem{STW}
Shawe-Taylor, J., Williamson, R.:
\newblock A {PAC} analysis of a {B}ayes estimator.
\newblock In: Proceedings of the Tenth Annual Conference on Computational
  Learning Theory, New York, ACM (1997)  2--9

\bibitem{McAllester}
McAllester, D.A.:
\newblock Some pac-bayesian theorems.
\newblock In: Proceedings of the Eleventh Annual Conference on Computational
  Learning Theory (Madison, WI, 1998), ACM (1998)  230--234

\bibitem{Catoni2004}
Catoni, O.:
\newblock Statistical Learning Theory and Stochastic Optimization.
\newblock Springer Lecture Notes in Mathematics (2004)

\bibitem{Catoni2007}
Catoni, O.:
\newblock {PAC-B}ayesian Supervised Classification (The Thermodynamics of
  Statistical Learning). Volume~56 of Lecture Notes-Monograph Series.
\newblock IMS (2007)

\bibitem{DT1}
Dalalyan, A.S., Tsybakov, A.B.:
\newblock Aggregation by exponential weighting, sharp {PAC}-{B}ayesian bounds
  and sparsity.
\newblock Machine Learning \textbf{72} (2008)  39--61

\bibitem{DT12a}
Dalalyan, A.S., Tsybakov, A.B.:
\newblock Sparse regression learning by aggregation and {L}angevin
  {M}onte-{C}arlo.
\newblock J. Comput. System Sci. \textbf{78} (2012)  1423--1443

\bibitem{DalSal}
Dalalyan, A.S., Salmon, J.:
\newblock Sharp oracle inequalities for aggregation of affine estimators.
\newblock The Annals of Statistics \textbf{40} (2012)  2327--2355

\bibitem{AL}
Alquier, P., Lounici, K.:
\newblock {PAC}-{B}ayesian bounds for sparse regression estimation with
  exponential weights.
\newblock Electronic Journal of Statistics \textbf{5} (2011)  127--145

\bibitem{AudibertCatoni}
Audibert, J.Y., Catoni, O.:
\newblock Robust linear least squares regression.
\newblock The Annals of Statistics \textbf{39} (2011)  2766--2794

\end{thebibliography}

\end{document}